\documentclass[10pt, a4paper]{article}
\usepackage{lrec2022} 
\usepackage{multibib}
\newcites{languageresource}{Language Resources}
\usepackage{graphicx}
\usepackage{tabularx}
\usepackage{soul}
\usepackage{booktabs}
\usepackage{multirow}
\usepackage{verbatim}	
\usepackage{xspace}
\usepackage[table]{xcolor}
\usepackage[colorinlistoftodos]{todonotes}
\usepackage{enumitem}
\newcommand*{\yoruba}{Yor\`ub\'a\xspace}

\usepackage{amssymb}  
\usepackage{pifont} 
\newcommand{\cmark}{\ding{51}}
\newcommand{\xmark}{\ding{55}}

\usepackage{pgfplots,pgfplotstable}
\pgfplotstableset{col sep=comma}
\pgfplotsset{compat=newest,}

\pgfplotsset{select coords between index/.style 2 args={
    x filter/.code={
        \ifnum\coordindex<#1\fi
        \ifnum\coordindex>#2\fi
    }
}}

\pgfplotstableread[col sep=comma]{results.csv}\FOnevsNoOfSent
\makeatletter
\pgfplotsset{
    /pgfplots/flexible xticklabels from table/.code n args={3}{%
        \pgfplotstableread[#3]{#1}\coordinate@table
        \pgfplotstablegetcolumn{#2}\of{\coordinate@table}\to\pgfplots@xticklabels
        \let\pgfplots@xticklabel=\pgfplots@user@ticklabel@list@x
    }
}
\makeatother

\usepackage{titlesec}
\titleformat{\section}{\normalfont\large\bfseries\center}{\thesection.}{1em}{}
\titleformat{\subsection}{\normalfont\SmallTitleFont\bfseries\raggedright}{\thesubsection.}{1em}{}
\titleformat{\subsubsection}{\normalfont\normalsize\bfseries\raggedright}{\thesubsubsection.}{1em}{}
\renewcommand\thesection{\arabic{section}}
\renewcommand\thesubsection{\thesection.\arabic{subsection}}
\renewcommand\thesubsubsection{\thesubsection.\arabic{subsubsection}}

\usepackage{epstopdf}
\usepackage[utf8]{inputenc}

\usepackage[T4,T1]{fontenc}
\usepackage[verbose]{newunicodechar}
\newenvironment{tfour}{\fontencoding{T4}\selectfont}{}

\usepackage{hyperref}
\usepackage{xstring}
\usepackage{lipsum}

\usepackage{color}
\usepackage{xcolor}

\usepackage{todonotes} 

\newcommand{\secref}[1]{\StrSubstitute{\getrefnumber{#1}}{.}{ }}

\newcommand{\naijasenti}{NaijaSenti\xspace}

\title{NaijaSenti: A Nigerian Twitter Sentiment Corpus for Multilingual Sentiment Analysis}

\name{Shamsuddeen Hassan Muhammad$^{1,2*+}$, David Ifeoluwa Adelani$^{3*}$, Sebastian Ruder$^{4}$\\ {\bf \large Ibrahim Sa'id Ahmad$^{5+}$, Idris Abdulmumin$^{6*+}$, Bello Shehu Bello$^{5+}$, Monojit Choudhury$^7$} \\ {\bf \large Chris Chinenye Emezue$^{8*}$,  Saheed Salahudeen Abdullahi$^{10+}$, Anuoluwapo Aremu$^{{11*}}$}\\
 {\bf \large Alípio Jorge$^{1,2}$ , Pavel Brazdil$^1$
}}

\address{$^{1}$ LIAAD - INESC TEC,  $^2$Faculty of Sciences-University of Porto, Portugal,  \\
	     $^3$Spoken Language Systems Group (LSV), Saarland University, Germany, $^4$Google Research
	     \\
      $^5$Faculty of Computer Science and Information Technology, Bayero University, Kano, Nigeria \\
	     $^6$Department of Computer Science, Ahmadu Bello University, Zaria, Nigeria, 
	     $^7$Microsoft Research India,\\
	     $^8$Technical University of Munich, Germany,
	     $^{9}$Clear Global, $^{10}$Kaduna state University\\
	      $^{*}$Masakhane NLP ,$^{+}$HausaNLP\\
	    \{\texttt{shmuhammad.csc, isahmad.it, bsbello.cs\}@buk.edu.ng}
}
\abstract{
Sentiment analysis is one of the most widely studied applications in NLP, but most work focuses on languages with large amounts of data. We introduce the first large-scale human-annotated Twitter sentiment dataset for the four most widely spoken languages in Nigeria---Hausa, Igbo, Nigerian-Pidgin, and \yoruba---consisting of around 30,000 annotated tweets per language, including a significant fraction of code-mixed tweets. We propose text collection, filtering, processing, and labeling methods that enable us to create datasets for these low-resource languages.
We evaluate a range of pre-trained models and transfer strategies on the dataset. We find that language-specific models and language-adaptive fine-tuning generally perform best. We release the datasets, trained models, sentiment lexicons, and code to incentivize research on sentiment analysis in under-represented languages.
\\ \newline  
}

\begin{document}

\maketitleabstract

\section{Introduction}

Sentiment analysis (SA) deals with the detection and classification of sentiment in texts~\cite{Pang2007OpinionMA}. In recent years, SA has attracted considerable interest, which can be attributed to its many vital applications.
However, most of the work on SA focuses on high-resource languages such as English \cite{yimam2020exploring} while languages with a limited amount of data remain poorly represented~\cite{nasim2020sentiment}. This problem is not unique to sentiment analysis, but affects NLP research as a whole \cite{joshi-etal-2020-state}. Recently, \newcite{nekoto2020participatory} and \newcite{adelani2021masakhaner} 
examined how socio-cultural factors hinder NLP for low-resource languages, potentially resulting in economic inequities \cite{weidinger2021ethical}.

With more than 200 million people and 522 native languages, Nigeria is the most populous and linguistically diverse country in Africa, as well as the third most multilingual country in the world.\footnote{\url{https://www.ethnologue.com/guides/countries-most-languages}} However, due to the lack of training data for many NLP applications, these languages are underserved by digital technology. Therefore, a concerted effort is required to create resources for such languages \cite{adelani2021masakhaner}.

In this paper, we present \emph{\naijasenti}\footnote{\url{https://github.com/hausanlp/NaijaSenti}}---an open-source  Twitter sentiment dataset for the four most spoken languages in Nigeria---Hausa, Igbo, Pidgin, and \yoruba. This is the largest labelled sentiment dataset in these languages to date. As the Twitter API does not support these languages, we propose methods to enable the collection, filtering, and annotation of such low-resource language data. Overall, we annotated around 30,000 tweets in Hausa, Igbo, \yoruba and Nigerian Pidgin (also known as Naija). The data highlight the challenges of sentiment analysis in these languages. For example, the absence of diacritics makes some tweets ambiguous in \yoruba and Igbo. In addition, code-mixing is a common occurrence, with about 43\% of Igbo tweets code-mixing between Igbo and English.

We conduct extensive experiments demonstrating that state-of-the-art pre-trained multilingual models achieve strong performance on sentiment classification on \naijasenti. The best models have been explicitly trained on unlabelled data in African languages during pre-training such as AfriBERTa \cite{ogueji-etal-2021-small} or using language-adaptive fine-tuning \cite{Pfeiffer2020mad-x}.



\newcolumntype{g}{>{\columncolor{Gray}}c}
\begin{table*}[ht]
 \begin{center}
 \rowcolors{1}{gray!8}{white}
 \resizebox{\textwidth}{!}{%
  \begin{tabular}{llcccc}
  \rowcolor{white}
    
    \toprule
    \textbf{Dataset} & \textbf{Language} & \textbf{Open-source} &  \textbf{Annotated/translated}  & \textbf{Code-mixed}  &  \textbf{Source}\\
    \midrule
    \newcite{Abubakar2021} & Hausa & \xmark & annotated &  \cmark &  Twitter  \\
     \newcite{ogbuju-onyesolu-2019-development}& Igbo & \xmark & translated & \xmark & General \\
     \newcite{umoh2020using} & Igbo & \xmark & annotated & \xmark & General\\ 
     \newcite{oyewusi2020semantic}* & Pidgin & \xmark & annotated/translated & \cmark  &  Twitter \\
     \newcite{orimaye2012sentiment} &  \yoruba &  \cmark &  annotated &  \cmark &  Youtube\\
    \newcite{iyanda2019predicting} & \yoruba & \xmark &     annotated & \xmark & General  \\
    \textbf{Ours} & Hausa, Igbo, \yoruba, Pidgin & \cmark & annotated  & \cmark & Twitter\\
     
    \bottomrule
    
  \end{tabular}
  }
  \caption{Summary of datasets used in six existing datasets on sentiment analysis in four major Nigerian languages in comparison to ours. *: The provided URL is no longer accessible.}
  \label{tab: onlystudiesinnative}
  \end{center}
\end{table*}
\paragraph{Contributions} 

The main contributions of this paper are: 
\begin{enumerate}[label={[\arabic*]}]

\item We curate large-scale manually annotated code-mixed and monolingual sentiment datasets for Hausa, Igbo, \yoruba and Pidgin languages.
\item We built a manually annotated sentiment lexicon in Hausa, Igbo, and \yoruba. We also semi-automatically develop translated emotion and sentiment lexicons in these languages.
\item We curate the largest Twitter corpus in each language that can be useful for other NLP downstream tasks.
\item We present several benchmark experiments on sentiment analysis in Hausa, Igbo, \yoruba, and Pidgin languages.   
\item We make the datasets and code freely available to foster further research in the NLP community. 

\end{enumerate}



\section{Related Work}


\paragraph{SA for low-resource languages} 

Sentiment analysis for low-resource languages has recently gained popularity  \cite{yimam2020exploring,xia2021metaxl,jovanoski2021sentiment} due to the availability of relatively large amounts of tweets in such languages. Several studies have investigated using Twitter for sentiment analysis\textemdash by either automatically building a Twitter corpus or manually annotating one. Notable studies that automatically built Twitter corpora include \newcite{go2009twitter}, \newcite{pak2010twitter}, and \newcite{wicaksono2014automatically}. More recently, \newcite{kwaik2020arabic} automatically built an Arabic Twitter sentiment analysis corpus using distant supervision and self-training. In contrast, other studies, such as ~\newcite{refaee2014arabic}, \newcite{brum2017building}, \newcite{mozetivc2016multilingual}, \newcite{nakov2019semeval2013}, and \newcite{moudjari2020algerian} employed native speakers or expert annotators to manually annotate the corpus. Our work is more closely related to \newcite{al2017arasenti} and \newcite{kwaik2020arabic}, as it involves both the use of emoji as a distantly supervised approach for tweet extraction and the use of a translated sentiment lexicon to filter tweets before manual annotation \cite{nakov2019semeval2013}

Despite advances in sentiment analysis for low-resource languages, indigenous Nigerian languages have received scant attention. This is mostly due to the absence of a freely accessible dataset in these languages. Nevertheless, there have been a significant number of studies on Nigerian code-mixed English \cite{nwofe2017pro,olaleye2018sentiment,oyebode2019social,kolajo2019sentiment,rakhmanov2020comparative,olagunju2020exploring,onyenwe2020impact,honkanen2021interjections}.  Most work on Nigerian languages has relied on automatically generated data, including the following:

\paragraph{Hausa} \newcite{Abubakar2021} built a Twitter corpus and introduced combined Hausa and English features in a classifier.

\paragraph{Igbo} \newcite{ogbuju-onyesolu-2019-development} 
translated an English sentiment lexicon \cite{hu2004mining} and manually added Igbo native words to create IgboSentiLex.~\newcite{umoh2020using} analysed Igbo emotion words using Interval Type-2 Fuzzy Logic.\paragraph{\yoruba}
\newcite{orimaye2012sentiment} built a \yoruba corpus from YouTube and applied a translated SentiWordNet for the sentiment analysis task. \newcite{iyanda2019predicting} created a  multi-domain corpus (health, business, education, politics) and used different classic ML classifiers such as SVM to predict sentiment in text.
\paragraph{Pidgin} \newcite{oyewusi2020semantic} built a Pidgin tweet corpus and used a translated VADER English lexicon for sentiment analysis.

Table~\ref{tab: onlystudiesinnative} summarizes the existing datasets for Nigerian languages; only two datasets are freely available, indicating that more work is needed to make indigenous datasets accessible and to stimulate research in these languages. To the best of our knowledge, ours is the first publicly available large-scale manually annotated dataset for sentiment analysis research in the following Nigerian languages: Hausa, Igbo, \yoruba and Nigerian Pidgin (see Appendix \secref{sec:language}~for the language description and characteristics.)

\section{Data Collection and Cleaning}




\begin{table*}[ht]
\begin{center}
\resizebox{\textwidth}{!}{%
\begin{tabular}{lp{60mm}p{60mm}l}

\toprule
\textbf{Language} & \textbf{Tweet} & \textbf{Translation into English}& \textbf{Sentiment}                                                   \\ \midrule
Hausa (\texttt{hau}) & {@}USER Aunt rahma i \textcolor{blue}{luv} u wallah irin totally dinnan & {@}USER Aunty rahma I swear I \textcolor{blue}{love} you very much  & positive \\
Igbo (\texttt{ibo})  & akowaro ya \textcolor{blue}{ofuma} nne kai \textcolor{blue}{daalu} nwanne mmadu we go dey alright las las & they told it \textcolor{blue}{well} my fellow sister \textcolor{blue}{well done} at the end we will be all right&positive   \\
Naija (\texttt{pcm})  & E don tay wey I don dey \textcolor{blue}{crush} on this \textcolor{blue}{fine} woman \ldots & I have had a \textcolor{blue}{crush} on the \textcolor{blue}{beautiful} woman for a while \ldots & positive \\
\yoruba (\texttt{yor}) & \textcolor{red}{on\'ir\`e\'egb\`e} al\'a\`ad\'ugb\`o ati \textcolor{red}{ol\'oj\'uk\`ok\`or\`o} & \textcolor{red}{mischievous} and \textcolor{red}{coveteous} neighbour & negative \\ \bottomrule
\end{tabular}
}
\caption{Examples of tweets, their English translation, and sentiment in different Nigerian languages. The Hausa and Igbo examples are code-switched with Naija. Sentiment-bearing words are highlighted in blue (positive) and red (negative).}
\label{tab:tweetsamples}
\end{center}
\end{table*}

\subsection{Data Collection}

Twitter provides easy access to a large amount of domain-independent and topic-independent public opinionated user-generated data. 
We collected tweets using the Twitter Academic API \footnote{\url{https://developer.twitter.com/en/products/twitter-api/academic-research}}, which
provides real-time and historical tweet data. The Twitter API supports retrieving tweets in 70 languages (including Amharic as the only African language) using language parameters. This makes it easy to extract a tweet in these languages. On the contrary, none of the languages considered in this work are supported by the API. Therefore, we consider different heuristic approaches to crawling tweets.



 
\paragraph{Stopwords, emoji, and sentiment words}




\newcite{caswell_language_2020} have shown that token-based filtering is a useful processing step for automatic language identification. Hence, we automatically built lists of common words (stopwords), which were verified by native speakers and used them to query the Twitter API to retrieve tweets in each language.~\newcite{go2009twitter} used emoticons and \newcite{kwaik2020arabic} used emojis as a distantly supervised approach to automatically classify subjective tweets as positive or negative. Using a similar approach, we used happy and sad emojis \cite{Kralj2015emojis} in combination with stopwords to query the Twitter API to extract tweets that contain stopwords and emojis. 
In addition, we used the Google Language API to translate the Affin lexicon \cite{nielsen2011new} into each of the languages (Hausa, Igbo, \yoruba), except Pidgin. We then filtered the tweets using the translated Affin sentiment lexicon to improve the likelihood of annotating sentiment-bearing tweets \cite{uzzaman2013semeval}.

\paragraph{Hashtags and Handles}

We used Twitter hashtags to crawl tweets from trending topics (e.g., \#Yorubaday) to collect sufficient tweets which are expected to be in the language under consideration. We also collect tweets from news handles (e.g., @bbchausa) which are expected to be factual and non-subjective. We selected the handles that tweet frequently in each language from the Indigenous Tweets\footnote{\url{http://indigenoustweets.com/}} website.

One downside of this approach is that Twitter conversations with a popular Twitter handle may dominate the dataset and may introduce a bias towards certain topics. For example, a Hausa Twitter conversation that involves the handle \texttt{@bbchausa} and another conversation involving the handle \texttt{@Rahmasadau} make up 54\% and 14\% respectively of collected tweets associated with Hausa handles. Limiting the number of tweets per conversation mitigates this problem. 

\subsection{Language Detection and Data Cleaning} 

Stopwords overlap across indigenous languages in a multilingual society such as Nigeria \cite{caswell_language_2020}. This results in tweets being collected in a language that differs from the query language. For example, using the stop word ``\emph{nke}'' to crawl tweets in Igbo produces tweets in Hausa, such as ``\emph{amin ya rabbi godiya nke}''. ~To mitigate this, we collected tweets based on locations where a language is predominantly spoken, using the location, longitude, latitude and radius parameters (25 miles) to specify a circular geographic area.

We also used Google CLD3\footnote{\url{https://github.com/google/cld3}} and Natural Language API\footnote{\url{https://cloud.google.com/natural-language/docs}} to detect the language of the collected tweets.  Pidgin is not supported by the API, so we used the stopword list to build an n-gram language detection tool to detect Pidgin. Before annotation, we cleaned the tweets. Retweets and duplicates were removed. We removed URLs and mentions, as well as trailing and redundant white spaces, converted all tweets to lowercase, and removed tweets with less than three words as they may contain insufficient information for sentiment analysis \cite{yang2018using}.

\section{Annotation and the NaijaSenti Dataset}



\subsection{Annotation Guidelines}

Our annotation guidelines focus on the classification of subjective tweets. A subjective tweet has a positive or negative emotion, opinion, or attitude \cite{refaee-rieser-2014-arabic}. 
We adapt a sentiment annotation guide from \cite{mohammad_practical_2016} and define five classes: positive (POS), negative (NEG), neutral (NEU), mixed (MIX) and indeterminate (IND). 

\paragraph{Positive (POS) Sentiment:} This occurs if a tweet implies positive sentiment, attitude and  emotional state. For example, a tweet implies a positive opinion or sentiment (e.g., \emph{“I love iPhone"}), positive emotional state (e.g., \emph{“we won the game last night"}), expression of support (e.g., \emph{“I will vote for PDP"}), thankfulness (e.g., \emph{“thank god she has not been kidnapped"}), success (e.g., \emph{“I passed all my exams"}), or positive attitude.

\paragraph{Negative (NEG) Sentiment:} This occurs if a tweet implies negative sentiment or emotion. For example, if a tweet implies negative sentiment (e.g., \emph{“ The iPhone camera is bad"}), negative emotional states such as failure, anger, and disappointment.

\paragraph{Neutral (NEU):} This occurs if the user’s tweet does not imply any positive or negative language directly or indirectly.  These are usually factual tweets, such as news.

\paragraph{Mixed (MIX):} This occurs if the user’s tweet implies both negativity and positivity  directly or indirectly.  For example, \emph{“I love an IPhone 10, but its camera is bad”}.

\paragraph{Indeterminate (IND):} This occurs if the users’ tweet does not fall into either positive, negative, neutral, and mixed, or if the annotator can only guess the class of a tweet, especially in the case of proverbs or sarcasm without sufficient context. We additionally use this class to label tweets in a different language (not code-mixed).






\subsection{Annotation Process}



\paragraph{Annotators training and preparation:}

For each language, we recruited three native speakers as annotators. The Annotators are both males and females between the ages of 20 and 45. We also recruited a coordinator for each language to supervise and ensure the quality of the annotation task. Annotators and coordinators have backgrounds in either computer science or linguistics and were trained on the annotation task using the LightTag annotation tool \cite{perry_lighttag:_2021}.

Data annotation is not a one-off process; it requires an agile approach with many iterations, collecting feedback from the annotators during the pilot stage, and refining the annotation guide to ensure that the annotators can achieve reasonable performance before moving to the next stage. We performed three iterations of the training and annotation practice of 100 tweets. For the first two iterations, the agreement among the annotators was poor. We asked the annotators for feedback and adapted a simplified sentiment questionnaire
annotation guide \cite{mohammad_practical_2016}.

\paragraph{Tweets annotation:}
The dataset was annotated in batches of 1,000 tweets by the annotators. For each batch, we adjudicated the cases in which the three annotators assigned a different label to a tweet. Annotators discuss these tweets, which allows them to address ambiguities, peculiar issues, and recommend ways to improve the annotation guidelines. We excluded these ambiguous tweets from the dataset. We iteratively update our annotation guide based on adjudication reports. Overall, the annotators annotated the following number of tweets: Hausa (35,000), Igbo (29,000), Pidgin (30,000) and \yoruba (33,000).

\paragraph{Determining the gold label:}
People often disagree on subjective concepts \cite{beddor_subjective_2019}. For example, person A, who has been using Apple products, says, ``The Apple iPhone camera is better than the Samsung camera'', while person B says, ``The Samsung camera is better''. This is an example of subjective disagreement in contrast to objective disagreement. Therefore, different from the simple majority vote approach \cite{davani2021dealing}, we introduced a new form of majority vote that involves an independent annotator who adjudicates subjective disagreement cases as follows:\footnote{We determine a single gold label for sentiment analysis in accordance with prior work. Future work may alternatively leverage annotator disagreement \cite{fornaciari-etal-2021-beyond}.}

\begin{itemize}
  \item Three-way agreement: Similar to the majority vote approach, if all three annotators agree on a label, we consider the agreed sentiment class to be the gold standard.
 
  \item Three-way disagreement: When all annotators disagree on a label, we discard the tweet.
   
  \item Two-way partial disagreement: If two of the annotators agree on a label, and the third annotator has a partial disagreement. For example, if two annotators classify a tweet as POS (or NEG), and the other annotator classifies it as a non-contradicting class such as NEU, we consider the POS (or NEG) classification to be the gold standard. 
  \item Two-way disagreement: If two of the annotators agree on a label, and the third annotator has a total disagreement. For example, if two annotators identify a tweet as POS and another as NEG or vice versa, the majority vote is not the final class (in this case, POS). To resolve such subjective disagreement, independent annotators review the disagreement and assign a final label.
\end{itemize}

\paragraph{Sentiment lexicons} We created sentiment lexicons in three languages (Hausa, Igbo, and \yoruba) based on NaijaSenti. We asked three annotators to tag words that convey negative or positive sentiment in a tweet. We used a simple majority vote. An independent annotator adjudicated cases where the annotators disagreed or only one person tagged a word as positive or negative. The distribution of the lexicon is presented in Table \ref{tab:stats}. We also created semi-automatically translated versions of the NRC emotion lexicon \cite{Mohammad13} and the AFFIN sentiment lexicon \cite{nielsen2011new} for Hausa, Igbo, and \yoruba. We used the Google Translate API\footnote{\url{https://cloud.google.com/translate}} to translate the lexicon. Afterwards, professional human translators verified and corrected the translations and added missing diacritics.




\subsection{Inter-Annotator Agreement}

We used the Fleiss kappa ($\kappa$) reliability measure~\cite{fleiss2013statistical} to determine the inter-annotator agreement (IAA) between the three annotators. The IAA for the 5-class and adjusted 3-class agreements are shown in Table~\ref{annot-stat-IAA}. The agreement between the five classes was not particularly high (e.g.,($\kappa$) = $0.35$) for Pidgin. However, according to the Fleiss classification~\cite{fleiss2013statistical}, an agreement greater than $0.40$ is considered reasonable (moderate) and beyond chance.

\begin{table}
\centering
\begin{tabular}{llrrrr}

\hline
&& \multicolumn{4}{r}{Sentiment datasets} \\
& sent. & hau & ibo & yor & pcm \\ \hline
\multirow{7}{*}{\rotatebox{90}{5-class}}& POS & $9,235$ & $5,621$ & $9,839$ &  $7,038$\\
& NEG & $9,033$& $4,726$& $5,003$ & $11,774$ \\
& NEU & $12,826$ & $14,877$& $14,356$& $2,205$\\
& IND &  $8$ & $1,909$ & $1,754$& $2,651$\\
& MIX & $1,466$& $19$ &  $622$ &  $1,696$\\ \cline{2-6}
\multicolumn{2}{r}{\textbf{Total}} & $32,568$& $27,152$& $31,574$ &  $29,837$\\
\multicolumn{2}{r}{IAA ($\kappa$)} & $0.487$& $0.488$& $0.555$ & $0.347 $ \\ \hline
\multirow{5}{*}{\rotatebox{90}{3-class}}& POS & $8,019$ & $5,395$ & $9,391$ & $5,839$ \\
& NEG & $8,119$ & $4,513$ & $4,638$ & $9,400$ \\
& NEU & $11,122$ & $13,380$ & $13,367$ & $2,004$ \\ \cline{2-6}
\multicolumn{2}{r}{\textbf{Total}} & $27,260$ & $23,288$ & $27,396$ & $17,243$\\
\multicolumn{2}{r}{IAA ($\kappa$)} & $0.607$ & $0.516$ & $0.600$ & $0.457$ \\ \hline

\end{tabular}
\caption{3-class and 5-class annotation and inter-Annotator agreement.}
\label{annot-stat-IAA}

\end{table}

We further computed the IAA ($\kappa$) (see Table~\ref{tab:IAAClass}) of each class with other classes to determine which classes the annotators find confusing or difficult and frequently disagree. Table~\ref{tab:IAAClass} indicates that annotators generally have the lowest overall agreement in the MIXED class, which includes elements of both the positive and negative classes, and some annotators identify it as either negative or positive. This highlights the subtlety of annotating mixed sentiment on social media and is in contrast to reviews where the annotation of mixed sentiment is clearer \cite{potts-etal-2021-dynasent}. To address this, we introduced an adjusted 3-class IAA agreement.

\begin{table}[tt]
\centering
\begin{tabular}{@{}lcccc@{}}
\toprule
\multicolumn{5}{c}{corpus}     \\ \midrule
\textbf{Class} & \textbf{hau} & \textbf{ibo} & \textbf{yor} & \textbf{pcm} \\\midrule
POS &  $0.626$   &  $0.542$    & $0.626$   &   $0.347$   \\
NEG      &  $0.518$   &   $0.521$  &  $0.553$   &   $0.416$   \\
NEU      &    $0.442$  &    $0.404$  &   $0.491$  &  $3.130$  \\
MIX      &   $0.297$    &   $0.020$  &   $0.242$  &   $0.130$  \\
IND      &   $0.045$   &  $0.591$   &  $0.764$    &  $0.679$ \\ \bottomrule
\end{tabular}
\caption{Fleiss kappa agreement among each class}
\label{tab:IAAClass}
\end{table}

In the adjusted 3-class agreement, we considered only positive, negative, and neutral as valid sentiment classes. We selected only tweets that have at least two labels in the valid classes and discarded the rest. For the selected tweets, where the label between two annotators is valid and the third label is in the invalid sentiment (Indeterminate or Mixed), we changed the label to the agreed valid label. For instance, given three annotation labels of a tweet as POS, POS, MIX, the third label is changed to POS, whereas the annotation labels of POS, POS, NEU are left unchanged. Table~\ref{annot-stat-IAA} shows the final statistics of at least two agreed tweets of the various datasets after converting to the 3-class annotation, and their corresponding inter-annotator agreements (IAA) using the Fleiss' Kappa ($\kappa$) metric.

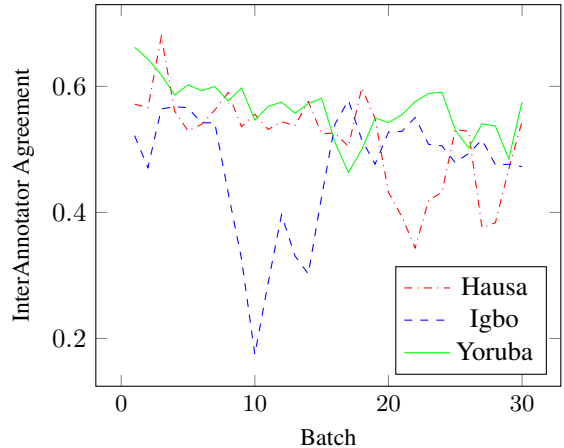
\begin{figure}[t]
\begin{center}
\begin{tikzpicture}
  \begin{axis}[
     width=\columnwidth,
     xlabel={\small Batch},
     ylabel={\small InterAnnotator Agreement},
     xticklabel style = {font=\footnotesize},
     legend pos=south east,
     legend entries={Hausa, Igbo, Yoruba},
  ]
  \addplot[dashdotted,red,mark=none] table[y index=1] {\IAAProgress};
  \addplot[dashed,blue,mark=none] table[y index=2] {\IAAProgress};
  \addplot[green,mark=none] table[y index=3] {\IAAProgress};
  \end{axis}
\end{tikzpicture}
\end{center}
\vspace{-2em}
\caption{Inter-annotators progress over thirty batches---one-thousand tweets per batch.}
\label{iaa_progress}
\end{figure}

To determine the performance of the IAA over time, Figure~\ref{iaa_progress} shows the IAA in three languages over 30 batches. We hypothesised that as the annotators became more experienced with the task, their annotation quality would improve. However, the overall performance of the IAA deteriorates over time. Igbo has the lowest performance drop. This suggests that familiarity with the task does not necessarily improve IAA. Only \yoruba annotators have some level of consistency that is not below $0.5$. Therefore, it is important to monitor the IAA as the annotation progresses and use some form of random quality check.

\subsection{Human Evaluations}

We assess human performance by re-annotating 200 random sample tweets by three different annotators \cite{warstadt2019neural,nangia2019human}. We take the majority vote as the final class. Human performance offers us an idea of the machine's upper bound performance and the reproducibility of the first three annotators \cite{warstadt2019neural}. Table \ref{tab:humaneval} shows the micro-F1 and Matthew's correlation coefficient (MCC) \cite{jurman2012comparison}. The human performance result validates the reliability of the corpus and is in line with previous literature \cite{rosenthal2017semeval}.

\pgfplotstableread[col sep=&, row sep=\\]{
Batch&	Hausa&	Igbo&	Yoruba\\
1&	0.5717&	0.5218&	0.662\\
2&	0.566&	0.4711&	0.6432\\
3&	0.6792&	0.5641&	0.6181\\
4&	0.5606&	0.5678&	0.5864\\
5&	0.5294&	0.566&	0.6024\\
6&	0.5396&	0.5421&	0.5932\\
7&	0.5624&	0.5423&	0.5996\\
8&	0.59&	0.4316&	0.5769\\
9&	0.5359&	0.3266&	0.5973\\
10&	0.5559&	0.1746&	0.5466\\
11&	0.5315&	0.2887&	0.5682\\
12&	0.5441&	0.3962&	0.575\\
13&	0.5376&	0.3311&	0.5574	\\
14&	0.5759&	0.3022&	0.5725	\\
15&	0.5246&	0.4253&	0.5812	\\
16&	0.5261&	0.5401&	0.5109	\\
17&	0.5046&	0.5779&	0.4634	\\
18&	0.5976&	0.5154&	0.4996	\\
19&	0.549&	0.4765&	0.5497	\\
20&	0.4317&	0.5273&	0.5424	\\
21&	0.3934&	0.5286&	0.555	\\
22&	0.3438&	0.5506&	0.5758	\\
23&	0.419&	0.508&	0.5886	\\
24&	0.4319&	0.5056&	0.5908	\\
25&	0.5315&	0.4789&	0.5308	\\
26&	0.5285&	0.4935&	0.5022	\\
27&	0.3755&	0.5152&	0.5404	\\
28&	0.3836&	0.4757&	0.5372 \\
29&	0.4686&	0.4764&	0.4848	\\
30&	0.5442&	0.4725&	0.5754	\\	
}\IAAProgress

\begin{table}[tt]
\centering
\begin{tabular}{@{}lcccc@{}}
\toprule
\textbf{Measures} & \textbf{hau} & \textbf{ibo} & \textbf{yor} & \textbf{pcm}\\\midrule
Micro-F1      &  $0.75$   &  $0.76$    & $0.85$   &  $0.78$ \\
MCC      &  $0.63$   &   $0.68$  &  $0.77$   &  $0.69$  \\
 \bottomrule
\end{tabular}
\caption{Human performance result using micro-F1 and Matthew's correlation coefficient.}
\label{tab:humaneval}
\end{table}

\subsection{NaijaSenti Statistics}

\begin{table*}[ht]
 \begin{center}
 \rowcolors{1}{gray!8}{white}
 \footnotesize
  \begin{tabular}{lccccccc}
  \rowcolor{white}
    
    \toprule
    \textbf{Languages} &     \textbf{mono-lingual} & \textbf{\#code-mixed} & 
    \textbf{token}  & 
    \textbf{Wordtype} & 
    \textbf{TTR}& 
    \textbf{neg words} & 
    \textbf{pos words}\\
    
    \midrule
    Hausa (\texttt{hau}) & $21,039$ &
    $6,426$ & $3,493,92$ & $ 30,747$ & 
    $0.09$ & $1,008$& $1,214$\\
    Igbo (\texttt{ibo})  & $8,688$  & $6,561$ & $1,830,02$ & $4,107 $ & $0.02$ & $1,180$ & $904$ \\
    Naija (\texttt{pcm}) & $-$ & $-$ & $3,669,68$ &$8,736$  & $0.06$& $-$&$-$ \\
    \yoruba (\texttt{yor}) & $18,662$ & $4,457$ & $40,897,6$ & $8,948$ & 0.02 & $2,185$ & $2,228$ \\
    
    \bottomrule
  \end{tabular}
  
  \caption{Key stats of NaijaSenti:    \#mono-lingual tweets, \#code-mix tweets, \#token, \#word types and type-to-token ratio (TTR)}

  \label{tab:stats}
  \end{center}
\end{table*}

Table \ref{annot-stat-IAA} shows the summary of our dataset with 5-class and adjusted 3-class. 
Other key statistical information, such as number of tokens, type of words, and type-token ration (TTR), which measure the lexical richness of a text are presented in Table~\ref{tab:stats}. We also show the number of monolingual and code-mixed tweets in each dataset. The percentage of code-mixed tweets highlights the highly multilingual setting in Nigeria. Code-mixing is more prevalent in Igbo (43\%) than in Hausa (23\%) and \yoruba (19\%). Code-mixing between English and a native language is more common than between native languages, but it can also occur between more than two native languages.

Hausa does not have diacritics and therefore has an insignificant number of indeterminate cases (only 8), unlike \yoruba and Igbo where the absence of diacritics may render a tweet incomprehensible and therefore lead to labelling it as indeterminate.~Pidgin has the highest number of indeterminate cases. This is because some tweets appear to be pidgin, but they are Nigerian English and, therefore, we consider them indeterminate. 

Tone in \yoruba 
helps to give meaning to words in context, especially words that have the same orthographic representation. For instance, the sentence ``Awon omo fo abo'' does not have a meaning without diacritics, and the annotators classify it as indeterminate (IND). However, the same sentence with diacritics can have two opposite meanings: \`{A}w{\d o}n {\d o}m{\d {\'{o}}} f{\d o} ab{\d {\'{o}}} (The children washed the dishes) has a positive meaning, and \`{A}w{\d o}n {\d o}m{\d {\'{o}}} f{\d {\'{o}}} ab{\d {\'{o}}} (The children broke the dishes) is negative.




Similarly, tonality is heavily used in Igbo. Many Twitter users do not write Igbo with diacritics. One reason is the lack of an Igbo keyboard that accepts and shows diacritics. Even if such a keyboard exists, it is not used by many. Although it may be fairly easy to understand the sentiment of Igbo tweets \emph{in context} on Twitter---either due to the presence of emojis or the context of the surrounding discourse, it is quite difficult and sometimes ambiguous to correctly annotate the tweets when they stand on their own. The example below highlights the impact of tone and punctuation marks on the same Igbo tweets but with different sentiment:



\begin{itemize}
    \item 
    \`{o} nw\`{e}kw\`{a}r\`{a} mgbe i naenwe sense \textbf{?} – Will you ever be able to talk sensibly? – You’re a fool.
    \item  \`{o} nw\`{e}kw\`{a}r\`{a} mgbe i naenwe sense – Sometimes you act with great maturity – I’m impressed.
    
    
\end{itemize}

Yes/No questions in Igbo are realized by a low tone on the subject pronoun, as in the first sentence above. So, with no tone and lacking punctuation, the author's intended meaning is difficult to determine.

\paragraph{Benchmark Data Split} To create a benchmark dataset, we use only three sentiment classes: negative, neutral, and positive. We split tweets in each class by 70\%, 10\% and 20\% ratios for the \texttt{TRAIN}, \texttt{DEV} and \texttt{TEST} splits as shown in Table \ref{tab:datasetsplit}. 


\begin{table}
\small
\centering
\begin{tabular}{@{}lcccc@{}}
\toprule
\textbf{lang.} & \textbf{TRAIN} & \textbf{DEV} & \textbf{TEST}  & \textbf{SPLIT} \\\midrule
hau & $ 18,989$  &  $2,714$     & $5,427$  & 70\slash10\slash20 \\
ibo      &  $12,930$    &   $1,84$  &  $3,697$   & 70\slash10\slash20 \\
pcm      &   $14,710$    &   $2,103$  &   $4,204$  & 70\slash10\slash20 \\
yor      &    $16,209$  &    $2,316$  &   $4,632$ &  70\slash10\slash20 \\
 \bottomrule
\end{tabular}
\caption{Benchmark data split}
\label{tab:datasetsplit}
\end{table}

\section{Experimental Setup}
\subsection{Sentiment Classification Models}
Sentiment classification is a well-studied problem in NLP and many machine learning models have been developed for this task. State-of-the-art approaches on English data use pre-trained language models (PLMs) such as BERT~\cite{devlin-etal-2019-bert} and RoBERTa~\cite{Liu2019RoBERTaAR}, which provide superior performance.
Multilingual variants of PLMs provide an opportunity to quickly adapt to various languages, including languages not seen during training \cite{Pfeiffer2020mad-x}. We compare several standard multilingual PLMs on the four languages. We fine-tune each model on the data of each language separately using the HuggingFace Transformer~\cite{wolf-etal-2020-transformers}. \autoref{sec:hyperparameter} provides the details of the hyper-parameters used for training. 

\paragraph{mBERT} is a multilingual variant of BERT pre-trained on 104 languages, including one Nigerian language---\yoruba. mBERT was pre-trained using masked language modeling (MLM) and next-sentence prediction task. We fine-tune the \texttt{mBERT-base-cased} model with 172M model parameters by adding a linear classification layer on top of the pre-trained transformer model.

\paragraph{XLM-R} Similar to mBERT, XLM-R~\cite{conneau-etal-2020-unsupervised} is a multilingual variant of RoBERTa pre-trained on 100 languages, including Hausa as the only Nigerian language. Unlike mBERT, XLM-R only uses MLM during pre-training. We use XLR-base with 270M model parameters for fine-tuning on the NaijaSenti corpus.  

\paragraph{RemBERT} scales up mBERT to a larger model size (559M) and decouples embeddings, which enables a larger output embedding size during pre-training, resulting in stronger pre-training and downstream performance \cite{chung2021rethinking}. RemBERT covers the three major Nigerian languages, except for Nigerian-Pidgin.

\paragraph{AfriBERTa} trains a RoBERTa-style model on 11 African languages~\cite{ogueji-etal-2021-small} including all four Nigerian languages in NaijaSenti. The model was trained on less than 1GB of data (since most African languages are low-resourced). We use AfriBERTa-large with 126M parameters. AfriBERTa has been shown to perform competitively on an African NER dataset~\cite{adelani2021masakhaner} despite its small model size and limited pre-training data.

\paragraph{mDeBERTaV3} Unlike the other four models pre-trained on the MLM task, mDeBERTaV3~\cite{He2021DeBERTaV3ID} makes use of ELECTRA-style~\cite{clark2020electra} pre-training where a discriminator is trained to detect replaced tokens instead of predicting masked tokens. mDEBERTaV3 does not support any of the Nigerian languages. We use the mDEBERTaV3-base model with 276M model parameters similar to XLM-R-base. 

\begin{table*}[ht]
 \begin{center}
 \resizebox{\textwidth}{!}{%
 \footnotesize
  \begin{tabular}{llcrrrrr}
    \toprule
    \textbf{Model} & \textbf{NG lang. supported} & \textbf{PLM size} &  \textbf{hau} & \textbf{ibo} & \textbf{pcm} & \textbf{yor} & \textbf{Avg}\\
    \midrule
    \multicolumn{2}{l}{\texttt{Majority Classifier}} \\
    Majority (Weighted F1) & -- & -- & $16.6$ & $26.9$ & $40.2$ & $19.0$ & $26.9$ \\
    Majority (Micro F1) & -- & -- & $33.3$ & $44.0$ & $56.0$ & $35.9$ & $43.4$ \\
    \midrule
    \multicolumn{2}{l}{\texttt{Multilingual PLMs}} \\
    AfriBERTa-large & hau, ibo, pcm, yor & 126M & $81.0^*_{\pm0.2}$ & $\mathbf{81.2^*_{\pm0.5}}$ & $70.9_{\pm0.7}$ & $80.2^*_{\pm0.6}$ & $\mathbf{78.3^*_{\pm0.3}}$ \\
    mBERT-base & yor & 172M & $77.8_{\pm0.5}$ & $79.8_{\pm0.5}$ & $69.0_{\pm0.2}$ & $77.6_{\pm0.9}$ & $76.9_{\pm0.3}$ \\
    XLM-R-base & hau & 270M & $78.4_{\pm1.0}$ & $79.9_{\pm0.7}$ & $\mathbf{73.3_{\pm0.3}}$ & $76.9_{\pm0.4}$ & $77.1_{\pm0.1}$ \\
    mDeBERTaV3-base & hau & 276M & $79.3_{\pm0.1}$ & $80.7_{\pm0.2}$ & $72.5^*_{\pm1.0}$ & $78.4_{\pm0.5}$ & $77.8_{\pm0.3}$ \\
    RemBERT & hau, ibo, yor & 559M & $79.0_{\pm0.7}$ & $79.9_{\pm0.4}$ & $\mathbf{73.3^*_{\pm1.4}}$ & $78.0_{\pm0.6}$ & $77.5_{\pm0.2}$ \\
    \midrule
    \multicolumn{2}{l}{\texttt{Multilingual PLMs+LAFT}} \\
    mBERT+LAFT (General) & hau / ibo / pcm / yor & 172M & $80.8_{\pm0.3}$ & $80.4_{\pm0.4}$ & $70.4_{\pm0.5}$ & $80.8_{\pm0.5}$ & $78.1_{\pm0.3}$ \\
    mBERT+LAFT (Tweet) & hau / ibo / pcm / yor & 172M & $79.3_{\pm0.6}$ & $77.7_{\pm0.6}$ & $70.7_{\pm0.7}$ & $76.8_{\pm0.3}$ & $76.1_{\pm0.2}$ \\
    XLM-R-base+LAFT (General) & hau / ibo / pcm / yor & 270M & $\mathbf{81.5^*_{\pm0.7}}$ & $80.8^*_{\pm0.8}$ & $70.0_{\pm1.1}$ & $\mathbf{80.9^*_{\pm0.4}}$ & $\mathbf{78.3^*_{\pm0.4}}$ \\
    XLM-R-base+LAFT (Tweet) & hau / ibo / pcm / yor & 270M & $79.5_{\pm0.9}$ & $77.0_{\pm0.5}$ & $71.1_{\pm1.3}$ & $76.2_{\pm0.4}$ & $75.9_{\pm0.2}$ \\
    \midrule
    \multicolumn{2}{l}{\texttt{Multi-task Multilingual PLMs}} \\
    AfriBERTa-large & hau, ibo, pcm, yor & 126M & $81.2^*_{\pm0.1}$ & $80.6^*_{\pm0.3}$ & $70.9_{\pm0.8}$ & $80.5^*_{\pm0.5}$ & $\mathbf{78.3_{\pm0.3}}$ \\
    mDeBERTaV3-base & hau & 276M & $79.4_{\pm0.4}$ & $79.6_{\pm0.2}$ & $72.7_{\pm0.4}$ & $78.4_{\pm0.2}$ & $77.5_{\pm0.1}$ \\
    \bottomrule
  \end{tabular}
}
  \caption{Weighted F1 evaluation of different Models. Average and standard deviation over 5 runs. Numbers with ``*'' are within the standard deviation of the best model. The models using language adaptive fine-tuning (LAFT) are trained on either the General domain or Twitter domain.}
  \label{tab:result}
  \end{center}
\end{table*}

\subsection{Language Adaptive Fine-tuning}

Many multilingual PLMs support only a few African languages. For example, mBERT only supports three African languages (Malagasy, Swahili, and \yoruba). Language adaptive fine-tuning (LAFT) is an effective method of adapting PLMs to a new language by fine-tuning PLMs MLM on unlabeled texts in the new language~\cite{Pfeiffer2020mad-x}. The approach is similar to domain-adaptive fine-tuning~\cite{Howard2018ulmfit,gururangan-etal-2020-dont}. LAFT has been shown to be very effective in improving NER performance in several African languages~\cite{alabi-etal-2020-massive,muller-etal-2021-unseen,adelani2021masakhaner}. To further improve the LAFT performance, we perform vocabulary augmentation using 99 most frequent wordpieces inspired by \cite{chau-etal-2020-parsing,pfeiffer-etal-2021-unks} before further pre-training the PLM. We experimented on two collections of monolingual data: (1) Twitter domain (often very small; less than 50K tweets for Igbo and \yoruba, and less than 600K tweets for Hausa and Nigerian-Pidgin), and (2) General domain (trained on mostly Common Crawl corpora, religious texts, and online news); for the latter, we use the checkpoints released by ~\cite{adelani2021masakhaner}.

\subsection{Multi-task Sentiment Classification}
In addition to fine-tuning separate models for each language, we trained a joint multi-task sentiment classification model on the four Nigerian languages by aggregating their training sets. The major advantage of this is that having a single model that can classify the sentiment in tweets in all major Nigerian languages facilitates deployment for practical applications. Knowledge from related languages may also be beneficial during transfer. This setting is possible because we are using multilingual PLMs that support multiple languages. 

\subsection{Cross-Lingual Transfer}
Lastly, we evaluate the zero-shot performance of a sentiment classifier trained on English tweets from SemEval-2017 Task 4 \cite{rosenthal2017semeval} on each of the four Nigerian languages. We also assess how many tweets from each of the Nigerian languages are needed to reach the zero-shot performance of a model transferred from English and to produce an accuracy score that is better than a majority classifier.

\section{Experimental Results}

\subsection{In-language Training}
\autoref{tab:result} shows the performance of several sentiment classification models for three-way sentiment classification on four Nigerian languages. As the corpora do not have a balanced number of samples for each label, we also computed a majority classifier based on the dominant label in the corpus. \texttt{hau}, \texttt{ibo} and \texttt{yor} have more \textit{neutral} tweets while \texttt{pcm} has more \textit{positive} tweets. The performance of the majority classifier using the weighted F1-score is around $16-45\%$ for all languages and $33-56\%$ using Micro F1-score. On the other hand, PLMs have a minimum F1-score of $70\%$, demonstrating their usefulness for sentiment analysis.

\paragraph{Multilingual PLMs} are quite similar in most cases with about a $1-3\%$ performance difference. The performance may depend on the language being seen during pre-training. mBERT has a slightly better performance ($+0.7\%$) for \texttt{yor} than XLM-R likely because \texttt{yor} was seen during pre-training. Similarly, XLM-R performs better for \texttt{hau}. RemBERT achieves slightly better performance than mBERT and XLM-R-base, demonstrating that a model with more capacity can improve performance. Surprisingly, we found mDeBERTaV3 that has only seen \texttt{hau} gives 
better results ($77.8\%$) than other models except for AfriBERTa that has been exclusively trained on African languages. 
mDeBERTaV3 makes use of replaced token detection (RTD), which has been shown to give superior performance for English~\cite{clark2020electra}. Overall, we found AfriBERTa to be the best baseline model for 
all languages because the model is more African language-centric. The main advantage of AfriBERTa is its smaller model size, which makes it easier to deploy especially on the African continent where most research labs cannot afford powerful GPUs. 

\paragraph{Language adaptive fine-tuning (LAFT)} has been shown to improve over the baseline with additional pre-training on monolingual data in the domain or language. \autoref{tab:result} shows some improvement over mBERT and XLM-R when we apply LAFT on the general domain, on average $2-3\%$ on \texttt{hau}, and \texttt{yor}, and $<1\%$ on \texttt{ibo}. For \texttt{pcm}, we only identified an improvement for mBERT ($+1.2\%$). Interestingly, applying LAFT on the Twitter domain did not improve performance. The main reason for this is the small size of the Twitter data. For example, \texttt{hau} was further pre-trained on CC100~\cite{conneau-etal-2020-unsupervised} corpus with over 318MB and 3 million sentences for the general domain, but for Twitter, we only have around 512K tweets (32MB), which are often short. In general, we found AfriBERTa to be competitive or better than LAFT for the Nigerian languages except for \texttt{pcm}. 

\paragraph{Multi-task sentiment classification} We trained a single model on all languages with minimal drop in performance. In this setting, we only trained on the best two multilingual PLMs: AfriBERTa and mDeBERTaV3. 
We observe only a slight drop in performance with mDeBERTa ($-0.3\%$) while the AfriBERTa performance is the same. This indicates that we could easily deploy a single sentiment classification model for the four major Nigerian languages, instead of multiple monolingual models.  

\begin{table}[t]
 \begin{center}
  \footnotesize
 \scalebox{0.9}{
  \begin{tabular}{llrrrrrr}
    \toprule
    \textbf{Model} &  \textbf{hau} & \textbf{ibo} & \textbf{pcm} & \textbf{yor} & \textbf{Avg}\\
    \midrule
    AfriBERTa-large & $\mathbf{58.4}$ & $\mathbf{47.7}$ & $50.5$ & $\mathbf{43.1}$ & $\mathbf{49.9}$ \\
    mBERT-base & $31.0$ & $37.0$ & $50.8$ & $39.5$ & $39.6$ \\
    XLM-R-base & $38.4$ & $37.8$ & $56.3$ & $26.7$ & $39.8$ \\
    mDeBERTaV3-base & $50.1$ & $47.2$ & $\mathbf{57.7}$ & $36.4$ & $47.9$ \\
    RemBERT & $54.0$ & $45.4$ & $55.9$ & $30.2$ & $46.4$ \\
    \bottomrule
  \end{tabular}
  }
  \caption{Transfer Learning experiments. PLMs are trained on English SemEval 2017 and evaluated on Nigerian languages in a zero-shot setting}
  \label{tab:transf_learn}
  \end{center}
\end{table}

\subsection{Zero-shot Cross-Lingual Transfer}

\autoref{tab:transf_learn} shows the results of zero-shot transfer from English SemEval 2017 Task 4 tweets to the four Nigerian languages. The English SemEval corpus consists of 11,763 tweets in the training set. \texttt{pcm} has the best zero-shot performance across all models because of its linguistic similarity to English, its lexifier language. Similarly, we found an impressive zero-shot performance for \texttt{hau} with at least $50.0\%$ F1-score when we train on AfriBERTa, mDeBERTaV3 and RemBERT. For \texttt{ibo}, the performance is over $45.4\%$ on the three best PLMs while the zero-shot evaluation for \texttt{yor} is slightly lower ($36-43\%$). 
AfriBERTa gave the best overall result in the zero-shot transfer, and it is significantly better than the majority classifier (weighted average) for all languages: \texttt{hau}, \texttt{ibo}, \texttt{pcm}, and \texttt{yor} are better by $41.8\%$, $20.8\%$, $4\%$, and $19.1\%$ respectively.

\subsection{Sample Efficiency in Transfer}
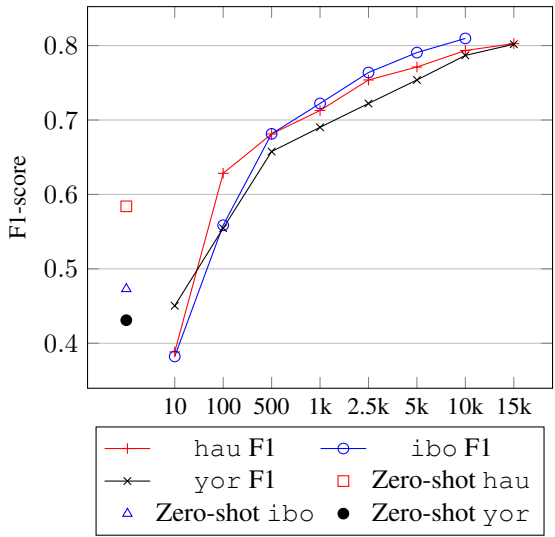
\begin{figure}[t]
\begin{center}
\begin{tikzpicture}
  \begin{axis}[
     width=\columnwidth,
     xlabel={\small Number of Sentences},
     ylabel={\small F1-score},
     flexible xticklabels from table={results.csv}{Number of Sentences}{col sep=comma},
     ymajorgrids=true,
     xtick=data,
     xticklabels={10,100,500,1k,2.5k,5k,10k,15k},
     xticklabel style = {font=\footnotesize},
     legend columns=2,
     legend style={at={(0.5,-0.1)},anchor=north},
     legend entries={\texttt{hau} F1, \texttt{ibo} F1, \texttt{yor} F1, Zero-shot \texttt{hau}, Zero-shot \texttt{ibo}, Zero-shot \texttt{yor}},
  ]
  \addplot[red,mark=+] table[select coords between index={1}{8}, x expr=\coordindex,y index=1] {\FOnevsNoOfSent};
  \addplot[blue,mark=o] table[select coords between index={1}{8}, x expr=\coordindex,y index=2] {\FOnevsNoOfSent};
  \addplot[mark=x] table[select coords between index={1}{8}, x expr=\coordindex,y index=3] {\FOnevsNoOfSent};
  \addlegendimage{red, only marks, mark=square};
  \addlegendimage{blue, only marks, mark=triangle};
  \addlegendimage{only marks, mark=*};
  \addplot[red,mark=square] table[select coords between index={0}{0}, x expr=\coordindex,y index=1] {\FOnevsNoOfSent};
  \addplot[blue,mark=triangle] table[select coords between index={0}{0}, x expr=\coordindex,y index=2] {\FOnevsNoOfSent};
  \addplot[mark=*] table[select coords between index={0}{0}, x expr=\coordindex,y index=3] {\FOnevsNoOfSent};
  \end{axis}
\end{tikzpicture}
\end{center}
\vspace{-2em}
\caption{Sample Efficiency on \texttt{hau}, \texttt{ibo}, and \texttt{yor} using the AfriBERTa model.}
\label{fig:sample_efficiency}
\end{figure}


\autoref{fig:sample_efficiency} shows the result of training a sentiment classification model with different numbers of samples (10, 100, 500, 1K, 2.5K, 5K, 10K, and 15K). We fine-tune AfriBERTa on \texttt{hau}, \texttt{ibo}, and \texttt{yor} datasets of different sizes. We observe an F1 score of $38-40\%$ with only 10 examples, which already exceeds the majority voting performance in \autoref{tab:result}. Surprisingly, with only 100 sentences, we exceed the zero-shot transfer performance from English language, and with at least 1000 sentences, we already reach a decent performance of $70\%$ F1. This result shows that we can leverage a multitask sentiment classification model trained on Nigerian languages to quickly adapt to other African languages with as few as 100 or 1000 annotated samples. Overall, we identify headroom for model improvement particularly in the zero-shot and few-shot cross-lingual transfer settings.

\section{Conclusions and Future Work}

In this paper, we present NaijaSenti---the first publicly available large-scale and manually annotated Twitter sentiment dataset for the four main Nigerian languages (Hausa, Igbo, Nigerian-pidgin, and \yoruba). We propose methods to enable the collection, filtering, and annotation of such low-resource language data. Additionally, we introduce a manually annotated sentiment lexicon in three languages (Hausa, Igbo, and \yoruba). We present benchmark experiments on Twitter sentiment dataset using state-of-the-art pre-trained language models and transfer learning. The results indicate that language-specific models and language-adaptive fine-tuning perform the best on average.  NaijaSenti has the potential to spark interest in sentiment analysis and other downstream NLP tasks in the languages involved. As future work, we plan to create benchmark experiments with our sentiment lexicon, and extend our dataset (NaijaSenti) to include other African languages (AfriSenti).

\section{Acknowledgements}

We thank Daan van Esch for feedback on a draft of this article. This work was carried out with support from Lacuna Fund, an initiative co-founded by The Rockefeller Foundation, Google.org, and Canada’s International Development Research Centre.~The views expressed herein do not necessarily represent those of Lacuna Fund, its Steering Committee, its funders, or Meridian Institute. This work is also partially funded by the National Funds through the Portuguese funding agency, FCT - Fundação para a Ciência e a Tecnologia, within project LA/P/0063/2020. We thank Tal Perry for providing the LightTag annotation tool. Finally, David Adelani acknowledges the EU-funded Horizon 2020 projects: COMPRISE (\texttt{http://www.compriseh2020.eu/}) under grant agreement No. 3081705 and ROXANNE under grant number 833635.


\section{Bibliographical References}\label{reference}

\bibliographystyle{NaijaSenti}
\bibliography{NaijaSenti}

\begin{thebibliography}{}

\bibitem[\protect\citename{Abubakar \bgroup et al.\egroup }2021]{Abubakar2021}
Abubakar, A.~I., Roko, A., Bui, A.~M., and Saidu, I.
\newblock (2021).
\newblock An enhanced feature acquisition for sentiment analysis of english and
  hausa tweets.
\newblock {\em International Journal of Advanced Computer Science and
  Applications}, 12(9).

\bibitem[\protect\citename{Adelani \bgroup et al.\egroup
  }2021]{adelani2021masakhaner}
Adelani, D.~I., Abbott, J., Neubig, G., D’souza, D., Kreutzer, J., Lignos,
  C., Palen-Michel, C., Buzaaba, H., Rijhwani, S., Ruder, S., Mayhew, S.,
  Azime, I.~A., Muhammad, S.~H., Emezue, C.~C., Nakatumba-Nabende, J., Ogayo,
  P., Anuoluwapo, A., Gitau, C., Mbaye, D., Alabi, J., Yimam, S.~M., Gwadabe,
  T.~R., Ezeani, I., Niyongabo, R.~A., Mukiibi, J., Otiende, V., Orife, I.,
  David, D., Ngom, S., Adewumi, T., Rayson, P., Adeyemi, M., Muriuki, G.,
  Anebi, E., Chukwuneke, C., Odu, N., Wairagala, E.~P., Oyerinde, S., Siro, C.,
  Bateesa, T.~S., Oloyede, T., Wambui, Y., Akinode, V., Nabagereka, D.,
  Katusiime, M., Awokoya, A., MBOUP, M., Gebreyohannes, D., Tilaye, H., Nwaike,
  K., Wolde, D., Faye, A., Sibanda, B., Ahia, O., Dossou, B. F.~P., Ogueji, K.,
  DIOP, T.~I., Diallo, A., Akinfaderin, A., Marengereke, T., and Osei, S.
\newblock (2021).
\newblock {MasakhaNER: Named Entity Recognition for African Languages}.
\newblock {\em Transactions of the Association for Computational Linguistics},
  9:1116--1131, 10.

\bibitem[\protect\citename{Al-Twairesh \bgroup et al.\egroup
  }2017]{al2017arasenti}
Al-Twairesh, N., Al-Khalifa, H., Al-Salman, A., and Al-Ohali, Y.
\newblock (2017).
\newblock Arasenti-tweet: A corpus for arabic sentiment analysis of saudi
  tweets.
\newblock {\em Procedia Computer Science}, 117:63--72.

\bibitem[\protect\citename{Alabi \bgroup et al.\egroup
  }2020]{alabi-etal-2020-massive}
Alabi, J., Amponsah-Kaakyire, K., Adelani, D., and Espa{\~n}a-Bonet, C.
\newblock (2020).
\newblock Massive vs. curated embeddings for low-resourced languages: the case
  of {Y}or{\`u}b{\'a} and {T}wi.
\newblock In {\em Proceedings of the 12th Language Resources and Evaluation
  Conference}, pages 2754--2762, Marseille, France, May. European Language
  Resources Association.

\bibitem[\protect\citename{Beddor}2019]{beddor_subjective_2019}
Beddor, B.
\newblock (2019).
\newblock Subjective {Disagreement}.
\newblock {\em Noûs}, 53(4):819--851, December.

\bibitem[\protect\citename{Brum and Nunes}2017]{brum2017building}
Brum, H.~B. and Nunes, M. d. G.~V.
\newblock (2017).
\newblock Building a sentiment corpus of tweets in brazilian portuguese.
\newblock {\em arXiv preprint arXiv:1712.08917}.

\bibitem[\protect\citename{Caswell \bgroup et al.\egroup
  }2020]{caswell_language_2020}
Caswell, I., Breiner, T., van Esch, D., and Bapna, A.
\newblock (2020).
\newblock Language {ID} in the {Wild}: {Unexpected} {Challenges} on the {Path}
  to a {Thousand}-{Language} {Web} {Text} {Corpus}.
\newblock In {\em Proceedings of the 28th {International} {Conference} on
  {Computational} {Linguistics}}, pages 6588--6608, Barcelona, Spain (Online).
  International Committee on Computational Linguistics.

\bibitem[\protect\citename{Chau \bgroup et al.\egroup
  }2020]{chau-etal-2020-parsing}
Chau, E.~C., Lin, L.~H., and Smith, N.~A.
\newblock (2020).
\newblock Parsing with multilingual {BERT}, a small corpus, and a small
  treebank.
\newblock In {\em Findings of the Association for Computational Linguistics:
  EMNLP 2020}, pages 1324--1334, Online, November. Association for
  Computational Linguistics.

\bibitem[\protect\citename{Chung \bgroup et al.\egroup
  }2021]{chung2021rethinking}
Chung, H.~W., Fevry, T., Tsai, H., Johnson, M., and Ruder, S.
\newblock (2021).
\newblock Rethinking embedding coupling in pre-trained language models.
\newblock In {\em International Conference on Learning Representations}.

\bibitem[\protect\citename{Clark \bgroup et al.\egroup }2020]{clark2020electra}
Clark, K., Luong, M.-T., Le, Q.~V., and Manning, C.~D.
\newblock (2020).
\newblock {ELECTRA}: Pre-training text encoders as discriminators rather than
  generators.
\newblock In {\em ICLR}.

\bibitem[\protect\citename{Conneau \bgroup et al.\egroup
  }2020]{conneau-etal-2020-unsupervised}
Conneau, A., Khandelwal, K., Goyal, N., Chaudhary, V., Wenzek, G., Guzm{\'a}n,
  F., Grave, E., Ott, M., Zettlemoyer, L., and Stoyanov, V.
\newblock (2020).
\newblock Unsupervised cross-lingual representation learning at scale.
\newblock In {\em Proceedings of the 58th Annual Meeting of the Association for
  Computational Linguistics}, pages 8440--8451, Online, July. Association for
  Computational Linguistics.

\bibitem[\protect\citename{Davani \bgroup et al.\egroup
  }2021]{davani2021dealing}
Davani, A.~M., D{\'\i}az, M., and Prabhakaran, V.
\newblock (2021).
\newblock Dealing with disagreements: Looking beyond the majority vote in
  subjective annotations.
\newblock {\em arXiv preprint arXiv:2110.05719}.

\bibitem[\protect\citename{Devlin \bgroup et al.\egroup
  }2019]{devlin-etal-2019-bert}
Devlin, J., Chang, M.-W., Lee, K., and Toutanova, K.
\newblock (2019).
\newblock {BERT}: Pre-training of deep bidirectional transformers for language
  understanding.
\newblock In {\em Proceedings of the 2019 Conference of the North {A}merican
  Chapter of the Association for Computational Linguistics: Human Language
  Technologies, Volume 1 (Long and Short Papers)}, pages 4171--4186,
  Minneapolis, Minnesota, June. Association for Computational Linguistics.

\bibitem[\protect\citename{Eberhard \bgroup et al.\egroup }2022]{ethnologue}
Eberhard, D.~M., Simons, G.~F., and (eds.), C. D.~F.
\newblock (2022).
\newblock Ethnologue: Languages of the world. twenty-third edition.

\bibitem[\protect\citename{Fleiss \bgroup et al.\egroup
  }2013]{fleiss2013statistical}
Fleiss, J.~L., Levin, B., and Paik, M.~C.
\newblock (2013).
\newblock {\em Statistical methods for rates and proportions}.
\newblock john wiley \& sons.

\bibitem[\protect\citename{$\forall$ \bgroup et al.\egroup
  }2020]{nekoto2020participatory}
$\forall$, ., Nekoto, W., Marivate, V., Matsila, T., Fasubaa, T., Fagbohungbe,
  T., Akinola, S.~O., Muhammad, S., Kabongo~Kabenamualu, S., Osei, S., Sackey,
  F., Niyongabo, R.~A., Macharm, R., Ogayo, P., Ahia, O., Berhe, M.~M.,
  Adeyemi, M., Mokgesi-Selinga, M., Okegbemi, L., Martinus, L., Tajudeen, K.,
  Degila, K., Ogueji, K., Siminyu, K., Kreutzer, J., Webster, J., Ali, J.~T.,
  Abbott, J., Orife, I., Ezeani, I., Dangana, I.~A., Kamper, H., Elsahar, H.,
  Duru, G., Kioko, G., Espoir, M., van Biljon, E., Whitenack, D., Onyefuluchi,
  C., Emezue, C.~C., Dossou, B. F.~P., Sibanda, B., Bassey, B., Olabiyi, A.,
  Ramkilowan, A., {\"O}ktem, A., Akinfaderin, A., and Bashir, A.
\newblock (2020).
\newblock Participatory research for low-resourced machine translation: A case
  study in {A}frican languages.
\newblock In {\em Findings of the Association for Computational Linguistics:
  EMNLP 2020}, Online.

\bibitem[\protect\citename{Fornaciari \bgroup et al.\egroup
  }2021]{fornaciari-etal-2021-beyond}
Fornaciari, T., Uma, A., Paun, S., Plank, B., Hovy, D., and Poesio, M.
\newblock (2021).
\newblock Beyond black {\&} white: Leveraging annotator disagreement via
  soft-label multi-task learning.
\newblock In {\em Proceedings of the 2021 Conference of the North American
  Chapter of the Association for Computational Linguistics: Human Language
  Technologies}, pages 2591--2597, Online, June. Association for Computational
  Linguistics.

\bibitem[\protect\citename{Go \bgroup et al.\egroup }2009]{go2009twitter}
Go, A., Bhayani, R., and Huang, L.
\newblock (2009).
\newblock Twitter sentiment classification using distant supervision.
\newblock {\em CS224N project report, Stanford}, 1(12):2009.

\bibitem[\protect\citename{Gururangan \bgroup et al.\egroup
  }2020]{gururangan-etal-2020-dont}
Gururangan, S., Marasovi{\'c}, A., Swayamdipta, S., Lo, K., Beltagy, I.,
  Downey, D., and Smith, N.~A.
\newblock (2020).
\newblock Don{'}t stop pretraining: Adapt language models to domains and tasks.
\newblock In {\em Proceedings of the 58th Annual Meeting of the Association for
  Computational Linguistics}, pages 8342--8360, Online, July. Association for
  Computational Linguistics.

\bibitem[\protect\citename{He \bgroup et al.\egroup }2021]{He2021DeBERTaV3ID}
He, P., Gao, J., and Chen, W.
\newblock (2021).
\newblock Debertav3: Improving deberta using electra-style pre-training with
  gradient-disentangled embedding sharing.
\newblock {\em ArXiv}, abs/2111.09543.

\bibitem[\protect\citename{Honkanen and
  M{\"u}ller}2021]{honkanen2021interjections}
Honkanen, M. and M{\"u}ller, J.
\newblock (2021).
\newblock Interjections and emojis in nigerian online communication.
\newblock {\em World Englishes}.

\bibitem[\protect\citename{Howard and Ruder}2018]{Howard2018ulmfit}
Howard, J. and Ruder, S.
\newblock (2018).
\newblock {Universal Language Model Fine-tuning for Text Classification}.
\newblock In {\em Proceedings of ACL 2018}.

\bibitem[\protect\citename{Hu and Liu}2004]{hu2004mining}
Hu, M. and Liu, B.
\newblock (2004).
\newblock Mining and summarizing customer reviews.
\newblock In {\em Proceedings of the tenth ACM SIGKDD international conference
  on Knowledge discovery and data mining}, pages 168--177.

\bibitem[\protect\citename{Iyanda and Abegunde}2019]{iyanda2019predicting}
Iyanda, A.~R. and Abegunde, O.
\newblock (2019).
\newblock Predicting sentiment in yor{\`u}b{\'a} written texts: A comparison of
  machine learning models.
\newblock In {\em Proceedings of SAI Intelligent Systems Conference}, pages
  416--431. Springer.

\bibitem[\protect\citename{Jaggar}2001]{Jaggar2001}
Jaggar, P.
\newblock (2001).
\newblock {\em Hausa}.
\newblock London Oriental and African language library. John Benjamins
  Publishing Company.

\bibitem[\protect\citename{Joshi \bgroup et al.\egroup
  }2020]{joshi-etal-2020-state}
Joshi, P., Santy, S., Budhiraja, A., Bali, K., and Choudhury, M.
\newblock (2020).
\newblock The state and fate of linguistic diversity and inclusion in the {NLP}
  world.
\newblock In {\em Proceedings of the 58th Annual Meeting of the Association for
  Computational Linguistics}, pages 6282--6293, Online, July. Association for
  Computational Linguistics.

\bibitem[\protect\citename{Jovanoski \bgroup et al.\egroup
  }2021]{jovanoski2021sentiment}
Jovanoski, D., Pachovski, V., and Nakov, P.
\newblock (2021).
\newblock Sentiment analysis in twitter for macedonian.
\newblock {\em arXiv preprint arXiv:2109.13725}.

\bibitem[\protect\citename{Jurman \bgroup et al.\egroup
  }2012]{jurman2012comparison}
Jurman, G., Riccadonna, S., and Furlanello, C.
\newblock (2012).
\newblock A comparison of mcc and cen error measures in multi-class prediction.
\newblock {\em PLOS ONE}, 7(8):1--8, 08.

\bibitem[\protect\citename{Kolajo \bgroup et al.\egroup
  }2019]{kolajo2019sentiment}
Kolajo, T., Daramola, O., and Adebiyi, A.
\newblock (2019).
\newblock Sentiment analysis on naija-tweets.
\newblock In {\em Proceedings of the 57th Annual Meeting of the Association for
  Computational Linguistics: Student Research Workshop}, pages 338--343.

\bibitem[\protect\citename{{Kralj Novak} \bgroup et al.\egroup
  }2015]{Kralj2015emojis}
{Kralj Novak}, P., Smailovi{\'c}, J., Sluban, B., and Mozeti\v{c}, I.
\newblock (2015).
\newblock Sentiment of emojis.
\newblock {\em PLoS ONE}, 10(12):e0144296.

\bibitem[\protect\citename{Kwaik \bgroup et al.\egroup }2020]{kwaik2020arabic}
Kwaik, K.~A., Chatzikyriakidis, S., Dobnik, S., Saad, M., and Johansson, R.
\newblock (2020).
\newblock An arabic tweets sentiment analysis dataset (atsad) using distant
  supervision and self training.
\newblock In {\em Proceedings of the 4th Workshop on Open-Source Arabic Corpora
  and Processing Tools, with a Shared Task on Offensive Language Detection},
  pages 1--8.

\bibitem[\protect\citename{Liu \bgroup et al.\egroup }2019]{Liu2019RoBERTaAR}
Liu, Y., Ott, M., Goyal, N., Du, J., Joshi, M., Chen, D., Levy, O., Lewis, M.,
  Zettlemoyer, L., and Stoyanov, V.
\newblock (2019).
\newblock Roberta: A robustly optimized bert pretraining approach.
\newblock {\em ArXiv}, abs/1907.11692.

\bibitem[\protect\citename{Mohammad and Turney}2013]{Mohammad13}
Mohammad, S.~M. and Turney, P.~D.
\newblock (2013).
\newblock Crowdsourcing a word-emotion association lexicon.
\newblock {\em Computational Intelligence}, 29(3):436--465.

\bibitem[\protect\citename{Mohammad}2016]{mohammad_practical_2016}
Mohammad, S.
\newblock (2016).
\newblock A {Practical} {Guide} to {Sentiment} {Annotation}: {Challenges} and
  {Solutions}.
\newblock In {\em Proceedings of the 7th {Workshop} on {Computational}
  {Approaches} to {Subjectivity}, {Sentiment} and {Social} {Media} {Analysis}},
  pages 174--179, San Diego, California. Association for Computational
  Linguistics.

\bibitem[\protect\citename{Moudjari \bgroup et al.\egroup
  }2020]{moudjari2020algerian}
Moudjari, L., Akli-Astouati, K., and Benamara, F.
\newblock (2020).
\newblock An algerian corpus and an annotation platform for opinion and emotion
  analysis.
\newblock In {\em Proceedings of The 12th Language Resources and Evaluation
  Conference}, pages 1202--1210.

\bibitem[\protect\citename{Mozeti{\v{c}} \bgroup et al.\egroup
  }2016]{mozetivc2016multilingual}
Mozeti{\v{c}}, I., Gr{\v{c}}ar, M., and Smailovi{\'c}, J.
\newblock (2016).
\newblock Multilingual twitter sentiment classification: The role of human
  annotators.
\newblock {\em PloS one}, 11(5):e0155036.

\bibitem[\protect\citename{Muller \bgroup et al.\egroup
  }2021]{muller-etal-2021-unseen}
Muller, B., Anastasopoulos, A., Sagot, B., and Seddah, D.
\newblock (2021).
\newblock When being unseen from m{BERT} is just the beginning: Handling new
  languages with multilingual language models.
\newblock In {\em Proceedings of the 2021 Conference of the North American
  Chapter of the Association for Computational Linguistics: Human Language
  Technologies}, pages 448--462, Online, June. Association for Computational
  Linguistics.

\bibitem[\protect\citename{Nakov \bgroup et al.\egroup
  }2019]{nakov2019semeval2013}
Nakov, P., Kozareva, Z., Ritter, A., Rosenthal, S., Stoyanov, V., and Wilson,
  T.
\newblock (2019).
\newblock Semeval-2013 task 2: Sentiment analysis in twitter.

\bibitem[\protect\citename{Nangia and Bowman}2019]{nangia2019human}
Nangia, N. and Bowman, S.~R.
\newblock (2019).
\newblock Human vs. muppet: A conservative estimate of human performance on the
  glue benchmark.
\newblock {\em arXiv preprint arXiv:1905.10425}.

\bibitem[\protect\citename{Nasim and Ghani}2020]{nasim2020sentiment}
Nasim, Z. and Ghani, S.
\newblock (2020).
\newblock Sentiment analysis on urdu tweets using markov chains.
\newblock {\em SN Computer Science}, 1(5):1--13.

\bibitem[\protect\citename{Nwofe}2017]{nwofe2017pro}
Nwofe, E.~S.
\newblock (2017).
\newblock Pro-biafran activists and the call for a referendum: A sentiment
  analysis of ‘biafraexit’on twitter after uk’s vote to leave the
  european union.
\newblock {\em Journal of Ethnic and Cultural Studies}, 4(1):65.

\bibitem[\protect\citename{Ogbuju and
  Onyesolu}2019]{ogbuju-onyesolu-2019-development}
Ogbuju, E. and Onyesolu, M.
\newblock (2019).
\newblock Development of a general purpose sentiment lexicon for {I}gbo
  language.
\newblock In {\em Proceedings of the 2019 Workshop on Widening NLP}, page~1,
  Florence, Italy, August. Association for Computational Linguistics.

\bibitem[\protect\citename{Ogueji \bgroup et al.\egroup
  }2021]{ogueji-etal-2021-small}
Ogueji, K., Zhu, Y., and Lin, J.
\newblock (2021).
\newblock Small data? no problem! exploring the viability of pretrained
  multilingual language models for low-resourced languages.
\newblock In {\em Proceedings of the 1st Workshop on Multilingual
  Representation Learning}, pages 116--126, Punta Cana, Dominican Republic,
  November. Association for Computational Linguistics.

\bibitem[\protect\citename{Ohiri-Aniche}2007]{onwu}
Ohiri-Aniche, C.
\newblock (2007).
\newblock Stemming the tide of centrifugal forces in {I}gbo orthography.
\newblock {\em Dialectical Anthropology}, 31(4):423--436.

\bibitem[\protect\citename{Olagunju \bgroup et al.\egroup
  }2020]{olagunju2020exploring}
Olagunju, T., Oyebode, O., and Orji, R.
\newblock (2020).
\newblock Exploring key issues affecting african mobile ecommerce applications
  using sentiment and thematic analysis.
\newblock {\em IEEE Access}, 8:114475--114486.

\bibitem[\protect\citename{Olaleye \bgroup et al.\egroup
  }2018]{olaleye2018sentiment}
Olaleye, S.~A., Sanusi, I.~T., and Salo, J.
\newblock (2018).
\newblock Sentiment analysis of social commerce: a harbinger of online
  reputation management.
\newblock {\em International Journal of Electronic Business}, 14(2):85--102.

\bibitem[\protect\citename{Onyenwe \bgroup et al.\egroup
  }2020]{onyenwe2020impact}
Onyenwe, I., Nwagbo, S., Mbeledogu, N., and Onyedinma, E.
\newblock (2020).
\newblock The impact of political party/candidate on the election results from
  a sentiment analysis perspective using\# anambradecides2017 tweets.
\newblock {\em Social Network Analysis and Mining}, 10(1):1--17.

\bibitem[\protect\citename{Orimaye \bgroup et al.\egroup
  }2012]{orimaye2012sentiment}
Orimaye, S.~O., Alhashmi, S.~M., and Eu-gene, S.
\newblock (2012).
\newblock Sentiment analysis amidst ambiguities in youtube comments on yoruba
  language (nollywood) movies.
\newblock In {\em Proceedings of the 21st International Conference on World
  Wide Web}, pages 583--584.

\bibitem[\protect\citename{Oyebode and Orji}2019]{oyebode2019social}
Oyebode, O. and Orji, R.
\newblock (2019).
\newblock Social media and sentiment analysis: The nigeria presidential
  election 2019.
\newblock In {\em 2019 IEEE 10th Annual Information Technology, Electronics and
  Mobile Communication Conference (IEMCON)}, pages 0140--0146. IEEE.

\bibitem[\protect\citename{Oyewusi \bgroup et al.\egroup
  }2020]{oyewusi2020semantic}
Oyewusi, W.~F., Adekanmbi, O., and Akinsande, O.
\newblock (2020).
\newblock Semantic enrichment of nigerian pidgin english for contextual
  sentiment classification.
\newblock {\em arXiv preprint arXiv:2003.12450}.

\bibitem[\protect\citename{Pak and Paroubek}2010]{pak2010twitter}
Pak, A. and Paroubek, P.
\newblock (2010).
\newblock Twitter as a corpus for sentiment analysis and opinion mining.
\newblock In {\em LREC}, volume~10, pages 1320--1326.

\bibitem[\protect\citename{Pang and Lee}2007]{Pang2007OpinionMA}
Pang, B. and Lee, L.
\newblock (2007).
\newblock Opinion mining and sentiment analysis.
\newblock {\em Found. Trends Inf. Retr.}, 2:1--135.

\bibitem[\protect\citename{Perry}2021]{perry_lighttag:_2021}
Perry, T.
\newblock (2021).
\newblock {LightTag}: {Text} {Annotation} {Platform}.
\newblock In {\em Proceedings of the 2021 {Conference} on {Empirical} {Methods}
  in {Natural} {Language} {Processing}: {System} {Demonstrations}}, pages
  20--27, Online and Punta Cana, Dominican Republic. Association for
  Computational Linguistics.

\bibitem[\protect\citename{Pfeiffer \bgroup et al.\egroup
  }2020]{Pfeiffer2020mad-x}
Pfeiffer, J., Vuli, I., Gurevych, I., and Ruder, S.
\newblock (2020).
\newblock {MAD-X: An Adapter-based Framework for Multi-task Cross-lingual
  Transfer}.
\newblock In {\em Proceedings of EMNLP 2020}.

\bibitem[\protect\citename{Pfeiffer \bgroup et al.\egroup
  }2021]{pfeiffer-etal-2021-unks}
Pfeiffer, J., Vuli{\'c}, I., Gurevych, I., and Ruder, S.
\newblock (2021).
\newblock {UNK}s everywhere: {A}dapting multilingual language models to new
  scripts.
\newblock In {\em Proceedings of the 2021 Conference on Empirical Methods in
  Natural Language Processing}, pages 10186--10203, Online and Punta Cana,
  Dominican Republic, November. Association for Computational Linguistics.

\bibitem[\protect\citename{Potts \bgroup et al.\egroup
  }2021]{potts-etal-2021-dynasent}
Potts, C., Wu, Z., Geiger, A., and Kiela, D.
\newblock (2021).
\newblock {D}yna{S}ent: A dynamic benchmark for sentiment analysis.
\newblock In {\em Proceedings of the 59th Annual Meeting of the Association for
  Computational Linguistics and the 11th International Joint Conference on
  Natural Language Processing (Volume 1: Long Papers)}, pages 2388--2404,
  Online, August. Association for Computational Linguistics.

\bibitem[\protect\citename{Rakhmanov}2020]{rakhmanov2020comparative}
Rakhmanov, O.
\newblock (2020).
\newblock A comparative study on vectorization and classification techniques in
  sentiment analysis to classify student-lecturer comments.
\newblock {\em Procedia Computer Science}, 178:194--204.

\bibitem[\protect\citename{Refaee and Rieser}2014a]{refaee2014arabic}
Refaee, E. and Rieser, V.
\newblock (2014a).
\newblock An arabic twitter corpus for subjectivity and sentiment analysis.
\newblock In {\em LREC}, pages 2268--2273.

\bibitem[\protect\citename{Refaee and Rieser}2014b]{refaee-rieser-2014-arabic}
Refaee, E. and Rieser, V.
\newblock (2014b).
\newblock An {A}rabic {T}witter corpus for subjectivity and sentiment analysis.
\newblock In {\em Proceedings of the Ninth International Conference on Language
  Resources and Evaluation ({LREC}'14)}, pages 2268--2273, Reykjavik, Iceland,
  May. European Language Resources Association (ELRA).

\bibitem[\protect\citename{Rosenthal \bgroup et al.\egroup
  }2017]{rosenthal2017semeval}
Rosenthal, S., Farra, N., and Nakov, P.
\newblock (2017).
\newblock Semeval-2017 task 4: Sentiment analysis in twitter.
\newblock In {\em Proceedings of the 11th international workshop on semantic
  evaluation (SemEval-2017)}, pages 502--518.

\bibitem[\protect\citename{Umoh \bgroup et al.\egroup }2020]{umoh2020using}
Umoh, U., Eyoh, I., Isong, E., Ekong, A., and Peter, S.
\newblock (2020).
\newblock Using interval type-2 fuzzy logic to analyze igbo emotion words.
\newblock {\em Journal of Fuzzy Extension and Applications}, 1(3):217--240.

\bibitem[\protect\citename{UzZaman \bgroup et al.\egroup
  }2013]{uzzaman2013semeval}
UzZaman, N., Llorens, H., Derczynski, L., Allen, J., Verhagen, M., and
  Pustejovsky, J.
\newblock (2013).
\newblock Semeval-2013 task 1: Tempeval-3: Evaluating time expressions, events,
  and temporal relations.
\newblock In {\em Second Joint Conference on Lexical and Computational
  Semantics (* SEM), Volume 2: Proceedings of the Seventh International
  Workshop on Semantic Evaluation (SemEval 2013)}, pages 1--9.

\bibitem[\protect\citename{Warstadt \bgroup et al.\egroup
  }2019]{warstadt2019neural}
Warstadt, A., Singh, A., and Bowman, S.~R.
\newblock (2019).
\newblock Neural network acceptability judgments.
\newblock {\em Transactions of the Association for Computational Linguistics},
  7:625--641.

\bibitem[\protect\citename{Weidinger \bgroup et al.\egroup
  }2021]{weidinger2021ethical}
Weidinger, L., Mellor, J., Rauh, M., Griffin, C., Uesato, J., Huang, P.-S.,
  Cheng, M., Glaese, M., Balle, B., Kasirzadeh, A., et~al.
\newblock (2021).
\newblock Ethical and social risks of harm from language models.
\newblock {\em arXiv preprint arXiv:2112.04359}.

\bibitem[\protect\citename{Wicaksono \bgroup et al.\egroup
  }2014]{wicaksono2014automatically}
Wicaksono, A.~F., Vania, C., Distiawan, B., and Adriani, M.
\newblock (2014).
\newblock Automatically building a corpus for sentiment analysis on indonesian
  tweets.
\newblock In {\em Proceedings of the 28th Pacific Asia Conference on Language,
  Information and Computing}, pages 185--194.

\bibitem[\protect\citename{Wolf \bgroup et al.\egroup
  }2020]{wolf-etal-2020-transformers}
Wolf, T., Debut, L., Sanh, V., Chaumond, J., Delangue, C., Moi, A., Cistac, P.,
  Rault, T., Louf, R., Funtowicz, M., Davison, J., Shleifer, S., von Platen,
  P., Ma, C., Jernite, Y., Plu, J., Xu, C., Le~Scao, T., Gugger, S., Drame, M.,
  Lhoest, Q., and Rush, A.
\newblock (2020).
\newblock Transformers: State-of-the-art natural language processing.
\newblock In {\em Proceedings of the 2020 Conference on Empirical Methods in
  Natural Language Processing: System Demonstrations}, pages 38--45, Online,
  October. Association for Computational Linguistics.

\bibitem[\protect\citename{Xia \bgroup et al.\egroup }2021]{xia2021metaxl}
Xia, M., Zheng, G., Mukherjee, S., Shokouhi, M., Neubig, G., and Awadallah,
  A.~H.
\newblock (2021).
\newblock Metaxl: Meta representation transformation for low-resource
  cross-lingual learning.
\newblock {\em arXiv preprint arXiv:2104.07908}.

\bibitem[\protect\citename{Yang \bgroup et al.\egroup }2018]{yang2018using}
Yang, X., Macdonald, C., and Ounis, I.
\newblock (2018).
\newblock Using word embeddings in twitter election classification.
\newblock {\em Information Retrieval Journal}, 21(2):183--207.

\bibitem[\protect\citename{Yimam \bgroup et al.\egroup
  }2020]{yimam2020exploring}
Yimam, S.~M., Alemayehu, H.~M., Ayele, A., and Biemann, C.
\newblock (2020).
\newblock Exploring amharic sentiment analysis from social media texts:
  Building annotation tools and classification models.
\newblock In {\em Proceedings of the 28th International Conference on
  Computational Linguistics}, pages 1048--1060.

\bibitem[\protect\citename{Årup Nielsen}2011]{nielsen2011new}
Årup Nielsen, F.
\newblock (2011).
\newblock A new anew: Evaluation of a word list for sentiment analysis in
  microblogs.

\end{thebibliography}

\appendix

\section*{Appendix}

\section{Overview of the Languages}
\label{sec:language}

With over 522 native languages, Nigeria is the most multilingual country in Africa and the third most multilingual country in the world.\footnote{\url{https://www.ethnologue.com/guides/countries-most-languages}} Although linguistically very diverse, the majority of the population speaks either Hausa, Igbo, \yoruba, or Nigerian-Pidgin. Therefore, our work focuses on these three indigenous Nigerian languages (Hausa, \yoruba, and Igbo) and Nigerian-Pidgin.

\paragraph{Hausa (\texttt{hau}):} Hausa is a Chadic (Afroasiatic) language that is spoken in 3 broad dialects\footnote{\url{https://www.mustgo.com/worldlanguages.com/hausa}}: Eastern, Western and Northern \cite{Jaggar2001}. Hausa is spoken by approximately 77 million people around the world, mostly in West Africa \cite{ethnologue}. The language is written in two different scripts: Ajami and the more common Boko script. The Boko script uses the Latin characters without p, q, v and x as well as the following additional letters: consonants  (\begin{tfour}\m{b}\end{tfour}, \begin{tfour}\m{d}\end{tfour}, \begin{tfour}\m{k}\end{tfour}, \begin{tfour}\m{y}\end{tfour}, kw, \begin{tfour}\m{k}\end{tfour}w, gw, ky, \begin{tfour}\m{k}\end{tfour}y, gy, sh, ts) and vowels (the long a, i, o, u, e and two additional diphthongs ai and au). Hausa is a tonal language with two tones: low and high, represented by the grave (e.g. ``\`{e}'') and acute (e.g. ``\'{e}'') accents respectively, which are usually not marked in everyday writing. The sentence structure follows the Subject-Verb-Object (SVO) syntax.

\paragraph{Igbo (\texttt{ibo}):} Igbo belongs to the Benue-Congo group of the Niger-Congo language family and is spoken by over 27 million people \cite{ethnologue}. It is the primary language of the Igbo people, an ethnic group of southeastern Nigeria, but is also spoken in some parts of Equatorial Guinea and Cameroon. There are approximately 30 Igbo dialects, some of which are not mutually intelligible. Igbo is written using the \d{O}nw\d{u} orthography \cite{onwu}. \d{O}nw\d{u} consists of 28 consonants and 8 vowels. Standard Igbo consists of eight vowels, and thirty consonants. Igbo is a tonal language.  Tone varies by dialect but in most dialects there are three main ones: high, low and downstep. A typical Igbo sentence follows subject-verb-object (SVO) order.

\paragraph{\yoruba (\texttt{yor}):} \yoruba belongs to the Yoruboid sub-branch of the Volta-Niger branch of the Niger-Congo language family. The language is spoken in the south-western parts of Nigeria stretching into some parts of Togo and Benin.  The \yoruba alphabet is based on the Latin script consisting of 18 consonants, 7 oral vowels, 5 nasal vowels and syllabic nasal consonants with additional characters like{\d e}, {\d o} , {\d s}, gb. The language uses tones: high, mid, and low tones. The Yoruba language is spoken by approximately 46 million people~\cite{ethnologue} 
, mostly in Nigeria, and Republic of Benin. 

\paragraph{Nigerian-Pidgin (\texttt{pcm}):} Nigerian-Pidgin, also known as Naija, is an English-based creole language spoken as a lingua franca across regions in Nigeria. It is rooted in the Krio of the English-based creole language family with an estimate of about 40M and 80M first and second language speakers respectively. Nigerian Pidgin uses the Latin script but has no standardised orthographic representation. The phonology of the language displays no suprasegmental features such as tone as in other African languages and it makes heavy usage of loan words from African and European languages.

\section{Model Hyper-parameters for
Reproducibility}
\label{sec:hyperparameter}
For the pre-trained models, we fine-tune the models using HuggingFace transformer tool ~\cite{wolf-etal-2020-transformers} with the batch size of $32$, maximum sequence length of $128$, number of epochs of $20$, and default learning rate ($5e-5$) for all models except for XLM-R and RemBERT where we set learning rate to be $2e-5$ to ensure model convergence. All the experiments were performed Nvidia V100 and RTX 2080 GPUs.

\end{document}